\documentclass[11pt]{article}
\usepackage[utf8]{inputenc}
\usepackage[final]{acl}
\usepackage{multirow, makecell}
\usepackage{amsmath}
\usepackage{algorithm}
\usepackage{algpseudocode}
\usepackage{amssymb}

\usepackage{times}
\usepackage{latexsym}
\usepackage{graphicx}

\usepackage{xcolor,colortbl}
\usepackage[T1]{fontenc}
\usepackage[utf8]{inputenc}
\usepackage{tcolorbox}
\tcbuselibrary{breakable}
\usepackage{mwe}
\usepackage{tabu}
\usepackage{booktabs}
\usepackage{subcaption}
\usepackage{microtype}
\usepackage{comment}
\usepackage[normalem]{ulem}
\useunder{\uline}{\ul}{}

\title{UoR-NCL at SemEval-2025 Task 1: Using Generative LLMs and CLIP Models for Multilingual Multimodal Idiomaticity Representation}

\author{
Thanet Markchom\textsuperscript{1} \and Tong Wu\textsuperscript{2}\and Liting Huang\textsuperscript{3} \and Huizhi Liang\textsuperscript{3} \\
\textsuperscript{1}Department of Computer Science, University of Reading, Reading, UK \\
\textsuperscript{2} Previously at School of Computing, Newcastle University, Newcastle upon Tyne, UK \\
\textsuperscript{3}School of Computing, Newcastle University, Newcastle upon Tyne, UK \\
\texttt{thanet.markchom@reading.ac.uk}, \texttt{tongwuwhitney@gmail.com} ,\\ \texttt{L.Huang29@newcastle.ac.uk},
\texttt{huizhi.liang@newcastle.ac.uk}
}

\begin{document}

\maketitle

\begin{abstract}

SemEval-2025 Task 1 focuses on ranking images based on their alignment with a given nominal compound that may carry idiomatic meaning in both English and Brazilian Portuguese. To address this challenge, this work uses generative large language models (LLMs) and multilingual CLIP models to enhance idiomatic compound representations. LLMs generate idiomatic meanings for potentially idiomatic compounds, enriching their semantic interpretation. These meanings are then encoded using multilingual CLIP models, serving as representations for image ranking. Contrastive learning and data augmentation techniques are applied to fine-tune these embeddings for improved performance.
Experimental results show that multimodal representations extracted through this method outperformed those based solely on the original nominal compounds. The fine-tuning approach shows promising outcomes but is less effective than using embeddings without fine-tuning. 

\end{abstract}

\section{Introduction}

In Natural Language Processing (NLP), generating representations for idiomatic expressions presents a significant challenge due to their inherent complexity and non-literal meanings~\cite{phelps-etal-2024-sign}. To address this challenge, SemEval-2025 Task 1: Advancing Multimodal Idiomaticity Representation (AdMIRe)~\cite{pickard2025semeval} introduced two subtasks: Subtask A and Subtask B. Subtask A involves ranking five images based on how well they represent the meaning of a potentially idiomatic nominal compound in a given context sentence, in both English and Brazilian Portuguese. This work focuses on Subtask A.

Existing NLP models, particularly those based on transformer architectures such as GPT~\cite{Radford2018ImprovingLU} and CLIP (Contrastive Language–Image Pre-training)~\cite{radford2021learningtransferablevisualmodels}, have made significant strides in language representation~\cite{markchom-etal-2022-uor,phelps-etal-2024-sign, xiong-etal-2024-ncl}. However, they often struggle with idiomatic expressions due to their reliance on surface-level word associations and compositional semantics~\cite{he2024enhancing}. This problem necessitates further exploration of methods that can improve the models' capacity to understand and represent idioms effectively.

To address this issue, this paper uses generative LLMs and multilingual CLIP models to tackle Subtask A in both English and Brazilian Portuguese. Specifically, an
LLM is used to produce idiomatic meanings for potentially idiomatic compounds. These generated meanings provide richer semantic information about the idiom and may better capture the compound's intended meaning compared to its original form. A multilingual CLIP model is then used to extract embeddings of the compounds (based on their generated meanings) and corresponding images to compute similarities and rank the images accordingly. Furthermore, to improve the effectiveness of the CLIP embeddings, the extracted embeddings are fine-tuned using a contrastive learning method combined with various data augmentation techniques (rotation, cropping, flipping, brightness and contrast adjustments, and Gaussian blur for images and back translation and paraphrasing for image captions). By combining generative LLMs and CLIP models, our approach offers a robust framework for generating more accurate idiomatic representations for this task.

\section{Proposed Method}

\begin{figure*}[ht]
    \centering
    \includegraphics[width=.88\linewidth]{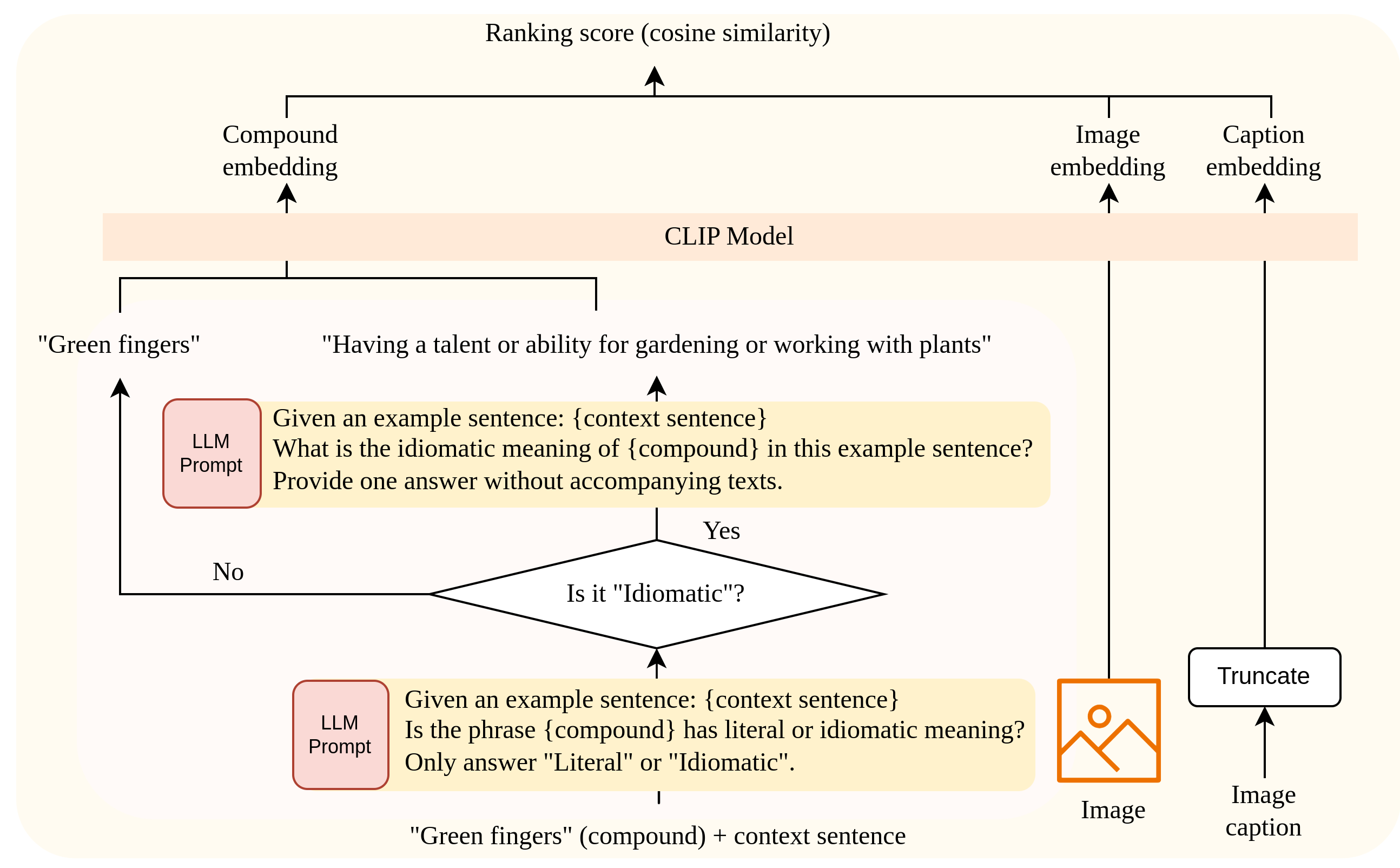}
    \caption{Overview of the proposed method: An LLM determines whether a compound is idiomatic or literal based on its context sentence. If idiomatic, the LLM generates its idiomatic meaning. A CLIP model then extracts embeddings of the original compound (if literal) or the generated meaning (if idiomatic), along with image and caption embeddings. Finally, cosine similarity is used to compute the ranking score.}
    \label{fig:diagram}
\end{figure*}

Figure \ref{fig:diagram} illustrates an overview of the proposed method. It starts with the idiomatic meaning generation step, where a generative LLM produces idiomatic meanings for potential idiomatic compounds. Next, the embedding extraction and image ranking step is described, where compound, image, and caption embeddings are extracted using the CLIP model and used to compute an image ranking score. Then, an ensemble method is introduced to enhance the accuracy of image ranking. Finally, a contrastive learning method to fine-tune the extracted CLIP embeddings is described.

\subsection{Idiomatic Meaning Generation}
\label{ref:detect_compound}

An LLM-based classification method is employed to determine whether a given compound phrase is used idiomatically or literally. The model is queried with a structured prompt that incorporates both the compound and its contextual sentence, as shown in Figure \ref{fig:diagram}. The LLM directly returns a classification label (``Idiomatic'' or ``Literal'') for each compound. To enhance classification robustness, this prompting process is repeated $T$ times, and the majority answer is selected for the final prediction. After obtaining the compound type, if it is classified as ``idiomatic'', the meaning of the compound is generated by prompting an LLM with the prompt shown in Figure \ref{fig:diagram}. This approach enables an automated method for generating idiomatic meanings.
    
    

\subsection{Embedding Extraction and Image Ranking}

In this step, the embeddings of the compound, candidate images, and their corresponding captions are extracted using a multilingual CLIP model. 
For the compound embedding, if its predicted type is ``literal'', the text embedding of the original compound, obtained from the CLIP model, is used as the compound embedding. If the type is ``idiomatic'', the text embedding of the generated idiomatic meaning from the previous step is used as the compound embedding. This ensures that, if the compound is idiomatic, its embedding (representation) incorporates additional information that reflects its idiomatic meaning. The same CLIP model is used to extract image embeddings for the candidate images. For caption embeddings, each caption is truncated at the end to the maximum input text length of the CLIP model, keeping the first part, and its text embedding is then extracted.
Once all embeddings are extracted, the ranking score $r_{c,i}$ of the nominal compound ($c$) and the candidate image $i$ is computed using the similarity between the compound embedding ($\mathbf{e}_c$) and each candidate image embedding ($\mathbf{e}_i$) along with its corresponding caption embedding ($\mathbf{e}_t$) as follows:
$r_{c,i} = s(\mathbf{e}_c, \mathbf{e}_i) + s(\mathbf{e}_c, \mathbf{e}_t)$
where $s(\cdot, \cdot)$  denotes a similarity function. This work uses cosine similarity to avoid magnitude invariance.

\subsection{Ensemble Method}

When generating idiomatic meanings for compounds, multiple LLMs can be utilized to capture diverse interpretations. To further enhance image ranking, an ensemble approach leveraging multiple LLMs is proposed. For each input (a compound, an image, and a caption), each LLM generates its interpretation of the compound's idiomatic meaning. A ranking score for the images is then computed based on these meanings. The individual scores from the LLMs are averaged to produce a final ranking score for each image. {There is no weighting, i.e., each LLM contributes equally to the final ranking score.
As for consistency, each model may interpret idiomatic meanings slightly differently and may not always be consistent with others. However, it is assumed that the majority of models will converge on the correct interpretation. By averaging their scores, individual biases are smoothed out, and commonly accurate interpretations are reinforced.} Overall, this ensemble method integrates insights from multiple LLMs, thereby improving the overall ranking performance.

\subsection{Fine-Tuning with Contrastive Learning}
\label{sec:finetune}

To enhance the CLIP embeddings and improve the alignment between idiomatic compounds and their corresponding images, fine-tuning is performed using a contrastive learning model.

\paragraph{Data Augmentation}
\label{sec:data_augmentation}
Data augmentation is applied to improve the robustness of the fine-tuning model. Images are randomly cropped to 450$\times$450 pixels (50\% probability), rotated within $\pm 45^{\circ}$ (50\% probability), and flipped horizontally (50\% probability) and vertically (50\% probability). Brightness and contrast are adjusted randomly (20\% probability), and Gaussian blur is applied (20\% probability) to simulate noise. For augmenting image captions, back translation and paraphrasing techniques are used. Back translation is performed using the Helsinki-NLP models—opus-mt-de-en and opus-mt-en-de—which translate the text from English to German and back to English \cite{tiedemann2023democratizing}. The google-t5/t5-base \cite{2020t5} model is used for paraphrasing.

\paragraph{Contrastive Learning Model}

To train the contrastive learning model, the dataset is prepared by constructing anchor-positive-negative triplets from the extracted embeddings. The compound embedding of each sample is an anchor. The ground-truth top-ranked image and its associated augmented image, caption, back-translated caption, and paraphrased caption are positive samples. Hard negatives are selected from the rest of the images and their associated augmented images, captions, back-translated captions, and paraphrased captions. Moreover, to enhance the learning process, soft negatives are randomly selected from other $K$ samples (other compounds) within the dataset.

The contrastive learning model is designed to project the embeddings into a shared latent space to maximize the similarity between anchor-positive pairs and minimize it for anchor-negative pairs. The model consists of a two-layer fully connected neural network with ReLU activation and dropout regularization. The output is projected into a latent space with a fixed dimensionality of $768$. The model is trained using the InfoNCE-based (Noise Contrastive Estimation) loss function \cite{oord2018representation} where the loss for each sample $s$ is  
\begin{equation}
     \mathcal{L}_s = - \frac{ \sum_{m=1}^{M} 
 \left[ \log \frac{f(\mathbf{a},\mathbf{p}_m)}{f(\mathbf{a},\mathbf{p}_m) + \sum_{n=1}^{N} f(\mathbf{a},\mathbf{n}_{m,n})
  } \right] }{M} 
\end{equation}
\normalsize
where $f(\mathbf{a},\mathbf{p}_m) = \exp \left( {s(\mathbf{a}, \mathbf{p}_m)}/{\tau} \right)$ and 
$f(\mathbf{a},\mathbf{n}_{m,n}) = \exp \left( {s(\mathbf{a}, \mathbf{n}_{m,n})}/{\tau}\right)$
where $M$ is the number of positive samples per anchor, $N$ is the number of negative samples per anchor, 
$\mathbf{a}$ is the anchor embedding, $\mathbf{p}_m$ is the positive sample embedding for modality $m$, $\mathbf{n}_{m,n}$ is the $n$-th negative sample embedding for modality $m$, $\tau$ is the temperature parameter, and $s(\cdot, \cdot)$ is the cosine similarity. The total loss is given by $\frac{1}{S}\sum_{s=1}^{S} \mathcal{L}_s$, where $S$ is the total number of training samples.

\section{Experimental Setup}

Three generative LLMs—\textbf{GPT-3.5}, \textbf{GPT-4}, and \textbf{GPT-4o}—were used for idiomatic meaning generation, and three multilingual CLIP models~\cite{carlsson-EtAl:2022:LREC}—\textbf{LABSE ViT-L/14 (LABSE)}, \textbf{XLM-R Large ViT-B/32 (XLM-32)}, and \textbf{XLM-R Large ViT-L/14 (XLM-14)}—for embedding generation. All methods in the experiments, including baselines and variations of the proposed method, are categorized as follows:
(1) \textbf{Baselines}: CLIP models applied directly to compounds to compute ranking scores without LLM-generated meanings;
(2) \textbf{Compound and Image without Fine-Tuning (CI)}: Ranking scores computed using only compound and image embeddings. Combinations of LLMs and CLIP models, including the ensemble method, were considered;
(3) \textbf{Compound, Image, and Caption without Fine-Tuning (CIC)}: Ranking scores computed using compound, image, and caption embeddings. Combinations of LLMs and CLIP models were the same as the previous approach;
(4) \textbf{Compound and Image with Fine-Tuning (CI-F)}: Ranking scores computed with fine-tuned compound and image embeddings (see Section \ref{sec:finetune}), using the best LLM and CLIP model combination from the non-fine-tuning approaches;
(5) \textbf{Compound, Image, and Caption with Fine-Tuning (CIC-F)}: Ranking scores computed with fine-tuned compound, image, and caption embeddings, using the best LLM and CLIP model combination as in the previous approach.

\paragraph{Datasets} Two datasets, English and Brazilian Portuguese, were provided for Subtask A of SemEval-2025 Task 1. The English dataset contains 70 training, 15 development, 15 test, and 100 extended test samples, while the Portuguese dataset contains 32 training, 10 development, 13 test, and 55 extended test samples.
Fine-tuning was performed on the augmented training sets (see Section \ref{sec:data_augmentation}). Note that the fine-tuning datasets are based on ground-truth compound types provided in the training sets. This avoids misclassification errors when using the proposed method for compound type prediction.
For each language, the augmented data was split into training (70\%), validation (10\%), and test (20\%) sets. This resulted in 50 training, 6 validation, and 14 test samples for English and 23 training, 3 validation, and 6 test samples for Brazilian Portuguese.


\paragraph{Hyperparameter Settings}
The number of repetitions for prompting the LLM to determine the compound type ($T$) was set to 5.
For \textbf{CI-F} and \textbf{CIC-F}, the hyperparameters for contrastive learning models were varied including batch size (16, 32), learning rate (1e-3, 1e-4, 1e-5), number of soft negatives $K$ (10, 30, 49), temperature $\tau$ (0.08, 0.09, 0.1), and dropout rate (0.1, 0.3, 0.5). The Adam optimizer was used. Early stopping was applied based on validation loss to prevent overfitting.

\paragraph{Evaluation Metrics}
For the compound-type prediction task, accuracy was used for evaluation. For the image ranking task, top-1 accuracy, Spearman's rank correlation and DCG score were used.

\section{Results and Discussion}

\subsection{Compound Type Detection Results}

Table \ref{tab:acc_NC_type} shows the accuracy of  GPT-3.5, GPT-4, and GPT-4o on the English and Portuguese training sets. From this table, GPT-4 outperformed the other models on both datasets. This highlights GPT-4's superior performance, which may be attributed to its more advanced architecture and training. GPT-4o also performed well on the English dataset but performed the worst on Portuguese. This lower performance of GPT-4o compared to GPT-4 could be due to the new tokenizer in GPT-4o. This tokenizer compresses tokens to reduce input length and improve efficiency \cite{openaiGPT4o}. Some word sequences that were previously tokenized as separate tokens in GPT-4 could be merged into a single token in GPT-4o, affecting the model's ability to understand a compound's meaning.

\begin{table}[]
\caption{Accuracy of compound type detection using different LLMs on English and Portuguese training sets}
    \centering
    \scalebox{.7}{
    \begin{tabular}{lll}
    \toprule
    {\textbf{Model}} & \textbf{English} & \textbf{Portuguese} \\
    \midrule
    GPT-3.5  & 0.7857  & 0.5938 \\
    GPT-4    & \textbf{0.8714}  & \textbf{0.6563} \\
    GPT-4o   & 0.8286  & 0.4688  \\ 
    \bottomrule
    \end{tabular}}
    
    \label{tab:acc_NC_type}
\end{table}

\subsection{Image Ranking Results}

Due to the small size of the development sets, only the results of the test and extended test sets are discussed in this section for a comprehensive evaluation. See Appendix \ref{apd:detailed_eval_results} for development set results.

Table \ref{tab:eval} shows the performance of baselines and variations of the proposed method on the complete test sets combining both the test and extended test samples. 
In this table, all the \textbf{baselines} performed worse than the proposed approach. This highlights the effectiveness of the proposed approach in generating more effective idiomaticity representations for the image ranking task.

\begin{table}[ht]
\caption{Evaluation results on the complete test sets (test and extended test sets combined) for both English (EN) and Brazilian Portuguese (PT). The highest values in each column are highlighted in bold.}
\centering
    \scalebox{.625}{
    \begin{tabular}{llrrrrrr}
    \toprule
      &  & \multicolumn{3}{c}{\textbf{Test EN}} & \multicolumn{3}{c}{\textbf{Test PT}} \\ 

      \cmidrule(lr){3-5} \cmidrule(lr){6-8}
      
    \multirow{-2}{*}{\bf LLM} & \multirow{-2}{*}{\bf CLIP model} & \multicolumn{1}{c}{\textbf{Acc}} & \multicolumn{1}{c}{\textbf{Corr}} & \multicolumn{1}{c}{\textbf{DCG}} & \multicolumn{1}{c}{\textbf{Acc}} & \multicolumn{1}{c}{\textbf{Corr}} & \multicolumn{1}{c}{\textbf{DCG}} \\

    \midrule
    
    \multicolumn{8}{l}{\cellcolor[HTML]{C0C0C0}Baselines} \\
    \textbf{-} & {\color[HTML]{000000} XLM-14} & {0.400} & {0.050} & {2.659} & {0.351} & {0.130} & {2.584} \\
    \textbf{-} & {\color[HTML]{000000} XLM-32} & {0.417} & {0.053} & {2.655} & {0.398} & {0.118} & {2.649} \\
    \textbf{-} & {\color[HTML]{000000} LABSE-14} & {0.409} & {0.126} & {2.648} & {0.445} & {0.161} & {2.666} \\
    
    \midrule
    
    \multicolumn{8}{l}{\cellcolor[HTML]{C0C0C0}Compound and Image without Fine-Tuning (CI)} \\
    GPT-3.5 & XLM-14 & 0.478 & 0.165 & 2.831 & 0.430 & 0.095 & 2.732 \\
    GPT-4 & XLM-14 & 0.504 & 0.126 & 2.906 & 0.418 & 0.157 & 2.749 \\
    GPT-4o & XLM-14 & 0.478 & 0.106 & 2.898 & 0.418 & 0.137 & 2.766 \\
    Ensemble & XLM-14 & 0.513 & 0.143 & 2.919 & 0.376 & 0.166 & 2.731 \\
    GPT-3.5 & XLM-32 & 0.435 & 0.102 & 2.757 & 0.487 & 0.107 & 2.823 \\
    GPT-4 & XLM-32 & 0.539 & 0.183 & 2.897 & 0.414 & 0.138 & 2.732 \\
    GPT-4o & XLM-32 & 0.513 & 0.171 & 2.899 & 0.481 & 0.172 & 2.829 \\
    Ensemble & XLM-32 & \textbf{0.557} & 0.122 & \textbf{2.939} & 0.450 & 0.175 & 2.798 \\
    GPT-3.5 & LABSE-14 & 0.470 & 0.177 & 2.816 & \textbf{0.530} & 0.184 & \textbf{2.846} \\
    GPT-4 & LABSE-14 & 0.496 & 0.163 & 2.883 & 0.471 & 0.178 & 2.778 \\
    GPT-4o & LABSE-14 & 0.504 & 0.187 & 2.899 & 0.481 & 0.194 & 2.825 \\
    Ensemble & LABSE-14 & 0.522 & \textbf{0.195} & 2.913 & 0.487 & \textbf{0.198} & 2.831 \\
    
    \midrule
    
    \multicolumn{8}{l}{\cellcolor[HTML]{C0C0C0}Compound, Image, and Caption without Fine-Tuning (CIC)} \\
    GPT-3.5 & XLM-14 & 0.287 & 0.043 & 2.480 & 0.315 & 0.005 & 2.503 \\
    GPT-4 & XLM-14 & 0.296 & 0.052 & 2.491 & 0.305 & 0.009 & 2.495 \\
    GPT-4o & XLM-14 & 0.296 & 0.063 & 2.573 & 0.315 & 0.023 & 2.530 \\
    Ensemble & XLM-14 & 0.287 & 0.061 & 2.509 & 0.293 & 0.071 & 2.490 \\
    GPT-3.5 & XLM-32 & 0.313 & 0.050 & 2.549 & 0.384 & 0.132 & 2.632 \\
    GPT-4 & XLM-32 & 0.357 & 0.074 & 2.594 & 0.368 & 0.107 & 2.623 \\
    GPT-4o & XLM-32 & 0.365 & 0.107 & 2.650 & 0.384 & 0.067 & 2.651 \\
    Ensemble & XLM-32 & 0.365 & 0.032 & 2.626 & 0.384 & 0.067 & 2.640 \\
    GPT-3.5 & LABSE-14 & 0.252 & 0.044 & 2.465 & 0.293 & 0.059 & 2.515 \\
    GPT-4 & LABSE-14 & 0.278 & 0.064 & 2.525 & 0.277 & 0.088 & 2.477 \\
    GPT-4o & LABSE-14 & 0.330 & 0.072 & 2.591 & 0.277 & 0.076 & 2.501 \\
    Ensemble & LABSE-14 & 0.278 & 0.066 & 2.525 & 0.293 & 0.089 & 2.501 \\ 
    
    \midrule
    
    \multicolumn{8}{l}{\cellcolor[HTML]{C0C0C0}Compound and Image with Fine-Tuning (CI-F)} \\
    GPT-3.5 & LABSE-14 & {0.391} & {0.027} & {2.709} & {-} & {-} & {-} \\
    GPT-4 & LABSE-14 & {0.400} & {0.079} & {2.778} & {-} & {-} & {-} \\
    GPT-4o & LABSE-14 & {0.365} & {0.056} & {2.707} & {-} & {-} & {-} \\
    
    \midrule
    
    \multicolumn{8}{l}{\cellcolor[HTML]{C0C0C0}Compound, Image, and Caption with Fine-Tuning (CIC-F)} \\
    GPT-3.5 & LABSE-14 & 0.391 & 0.053 & 2.697 & - & - & {-} \\
    GPT-4 & LABSE-14 & 0.417 & 0.155 & 2.813 & - & - & {-} \\
    GPT-4o & LABSE-14 & 0.374 & 0.084 & 2.722 & -  & - & {-} \\ 
    \bottomrule
    \end{tabular}}
    
    \label{tab:eval}
\end{table}

As for \textbf{CI}, the results show that the ensemble method with XLM-32 achieved the best top-1 accuracy and DCG score for English. For Portuguese, the method using GPT-3.5 with LABSE-14 performed the best in top-1 accuracy and DCG score. This suggests that these methods were particularly effective at selecting the most similar images that matched the compounds. In contrast, the ensemble method using LABSE-14 outperformed the others in terms of correlation for both languages. This suggests its potential for capturing nuanced levels of similarity between images and compounds.

Considering \textbf{CIC}, the methods in this approach overall performed worse compared to \textbf{CI}. This suggests that the addition of caption embeddings without fine-tuning did not significantly enhance the models' ability to match compounds with images effectively. One possible reason is that the captions are lengthy, making their embeddings from the CLIP models less effective.

\begin{figure}
    \centering
    \begin{subfigure}[b]{\linewidth}
         \includegraphics[width=\linewidth]{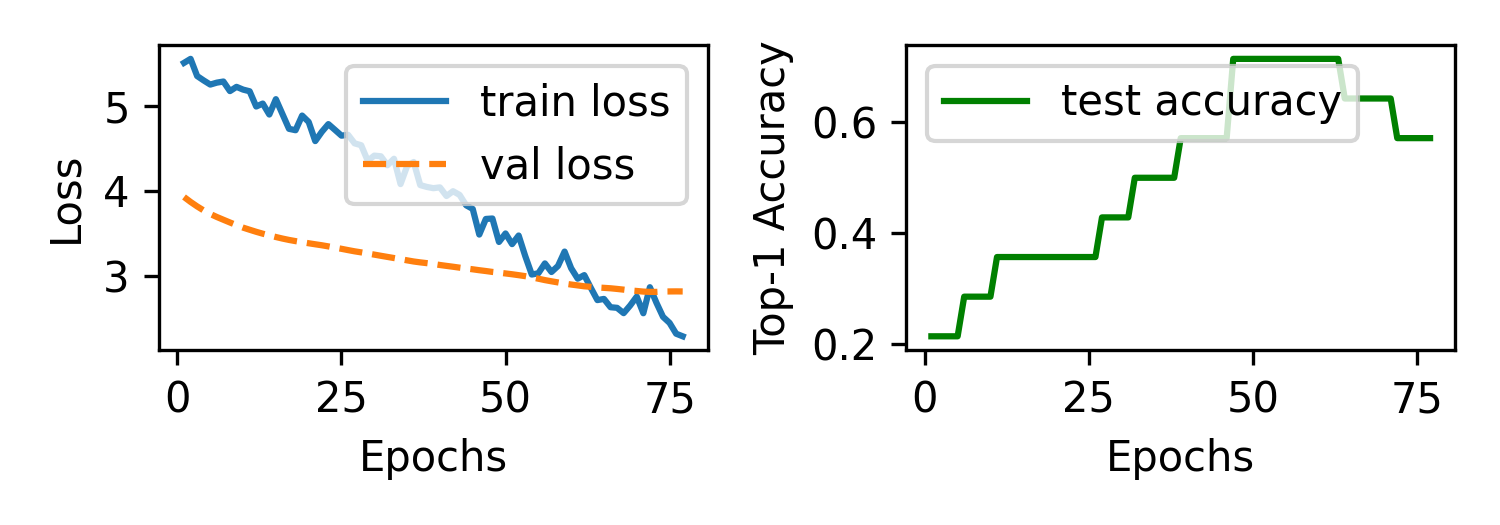} \vspace{-10mm}
         \caption{GPT-3.5}
     \end{subfigure} 
     
     \begin{subfigure}[b]{\linewidth}
         \includegraphics[width=\linewidth]{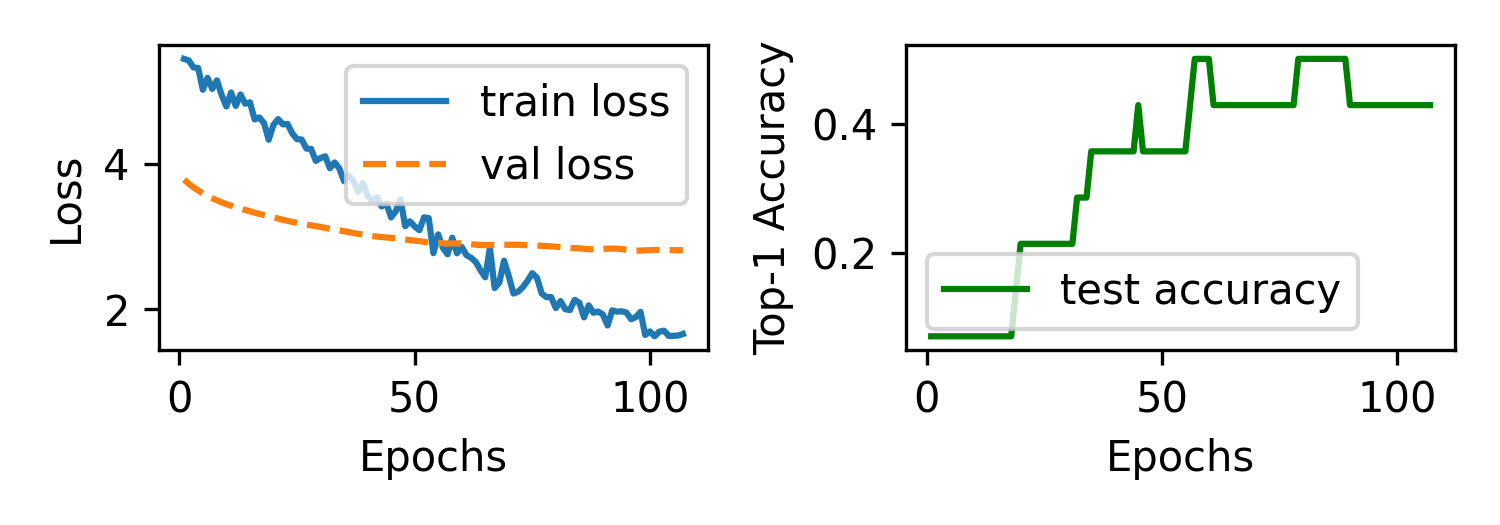}
         \vspace{-10mm}
         \caption{GPT-4}
     \end{subfigure}
     
     \begin{subfigure}[b]{\linewidth}
         \includegraphics[width=\linewidth]{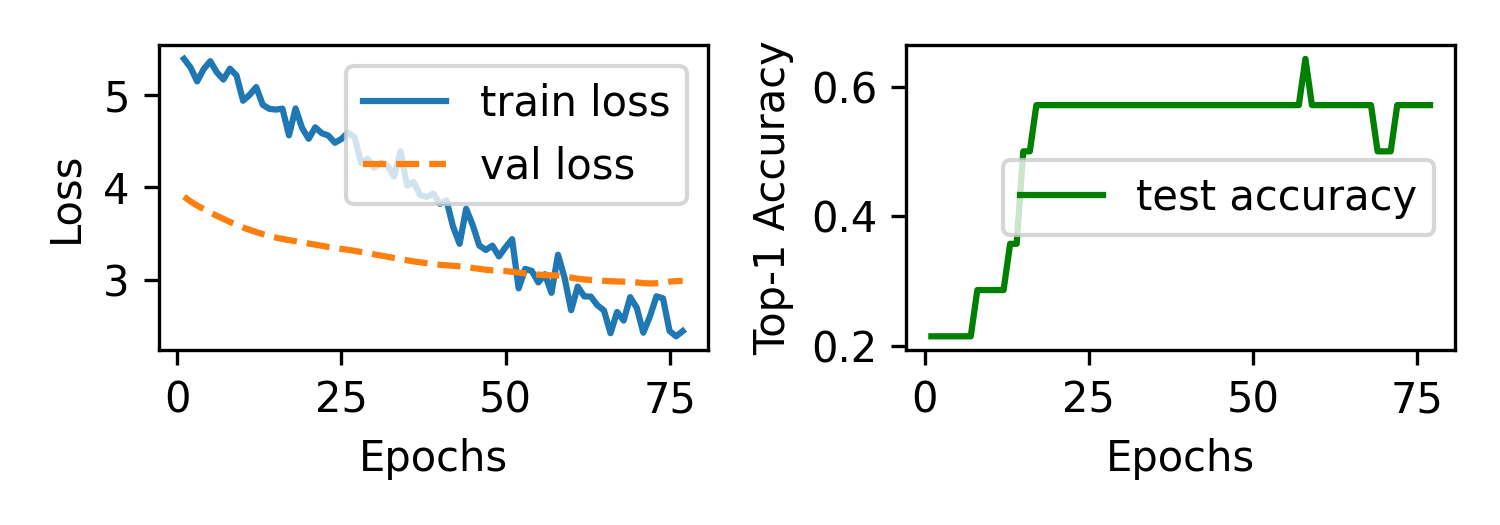}
         \vspace{-10mm}
         \caption{GPT-4o}
     \end{subfigure}

     \caption{Training loss, validation loss, and test accuracy, during the fine-tuning of embeddings obtained using LABSE-14, with GPT-3.5, GPT-4, and GPT-4o used for idiomatic meaning generation.}
     \label{fig:training_losses}
\end{figure}

Based on the results of \textbf{CI} and \textbf{CIC}, LABSE-14 demonstrated the highest effectiveness in ranking. Consequently, the embeddings obtained using LABSE-14 with different LLMs were fine-tuned in \textbf{CI-F} and \textbf{CIC-F}. Multiple contrastive models were trained on individual sets of embeddings from various LLMs. The selected hyperparameters for each model can be found in Appendix \ref{apd:hyperparams}. Overall, the fine-tuned embeddings did not perform as well as the non-fine-tuned embeddings. Figure \ref{fig:training_losses} shows the training and validation losses, as well as the test accuracy, during the fine-tuning of embeddings obtained using LABSE-14 with GPT-3.5, GPT-4, and GPT-4o. These figures suggest that the models effectively learned the fine-tuned embeddings, as test accuracy gradually increased over training epochs. However, the models began overfitting before the test accuracy could improve further. This could be due to the amount of training data being insufficient for the model to generalize well to unseen data. {Extra data augmentation could improve fine-tuning by introducing linguistic and visual diversity within existing, seen idioms. This may help the model more accurately match images in different styles to the generated meanings of seen idioms expressed with varying wordings. However, this may not be effective for unseen idioms if they share no common meanings with those in the training set. Similarly, the use of regularization may help generalize the model's ability to match images with seen idioms in the training set, but it might not improve generalization to unseen idioms.}
Due to the lack of performance improvement on the English dataset during fine-tuning, experiments on the Portuguese dataset were not conducted.

More detailed results on the individual test and extended test sets for both languages can be found in Appendix \ref{apd:detailed_eval_results} (Table \ref{tab:appendix_eval_results}).

\begin{figure}[h!]
    \centering
    \begin{subfigure}[b]{.44\linewidth}
         \includegraphics[width=\linewidth]{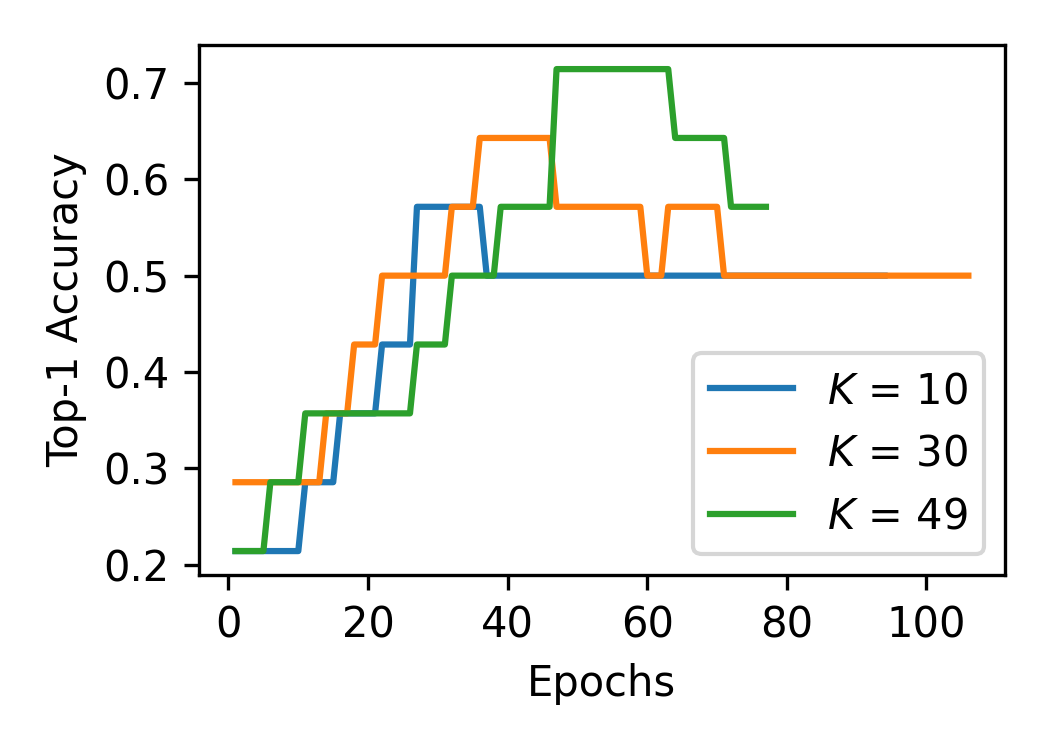} \caption{$K$}
     \end{subfigure}      
     \begin{subfigure}[b]{.44\linewidth}
         \includegraphics[width=\linewidth]{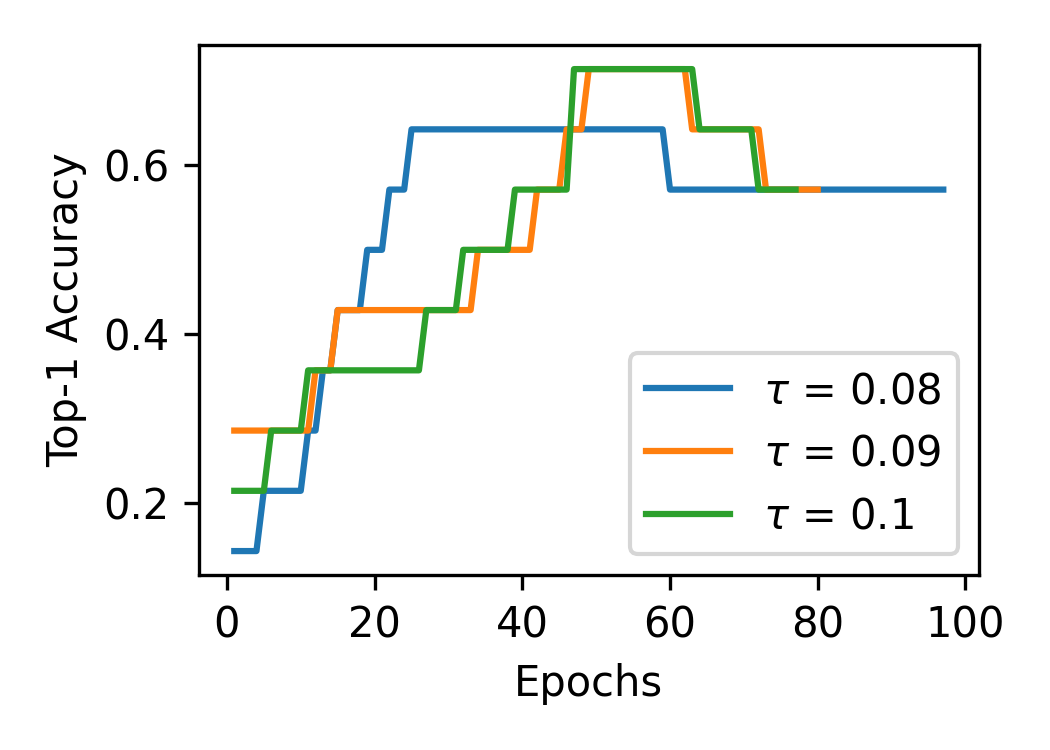}
         \caption{$\tau$}
     \end{subfigure} 
     \begin{subfigure}[b]{.44\linewidth}
         \includegraphics[width=\linewidth]{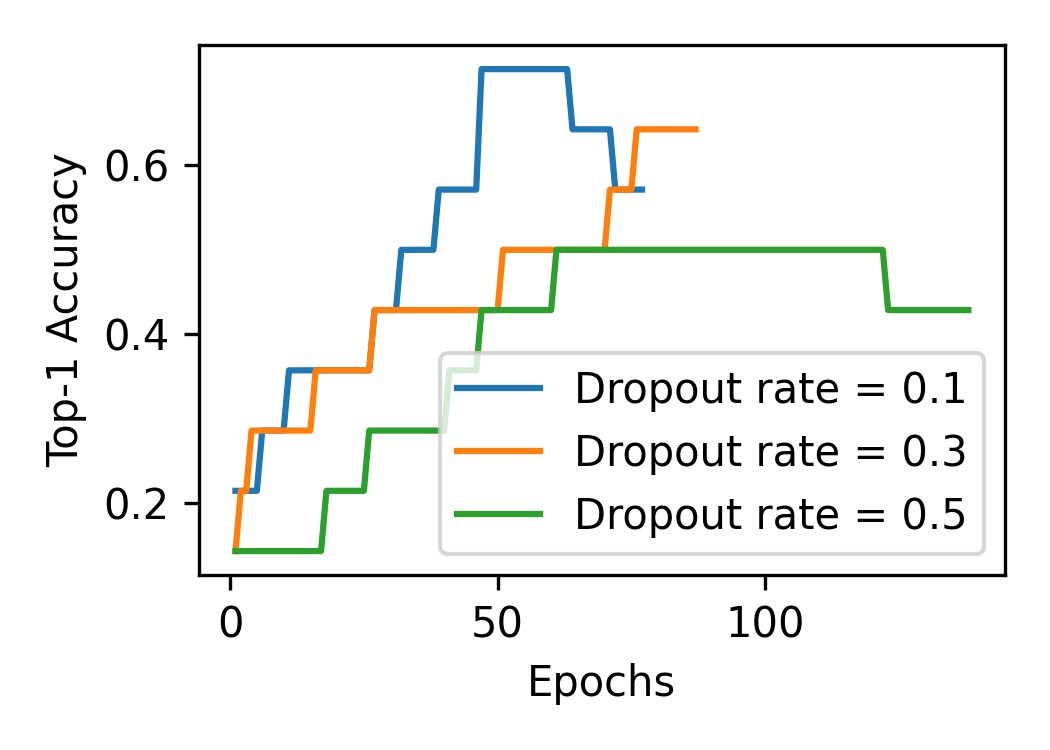}
         \caption{Dropout rate}
     \end{subfigure}
     \caption{Top-1 accuracy on the test set during the fine-tuning of embeddings obtained using LABSE-14, with GPT-3.5 used for idiomatic meaning generation.}
     \label{fig:params_GPT35}
    \centering
    \begin{subfigure}[b]{.44\linewidth}
         \includegraphics[width=\linewidth]{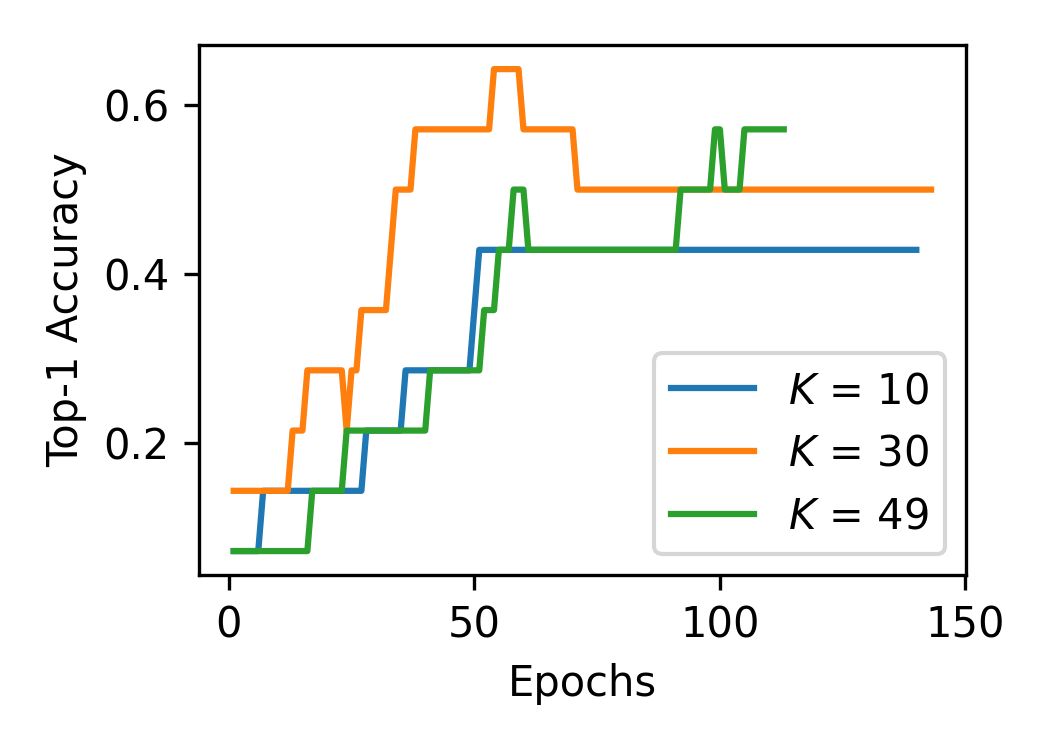} \caption{$K$}
     \end{subfigure}      
     \begin{subfigure}[b]{.44\linewidth}
         \includegraphics[width=\linewidth]{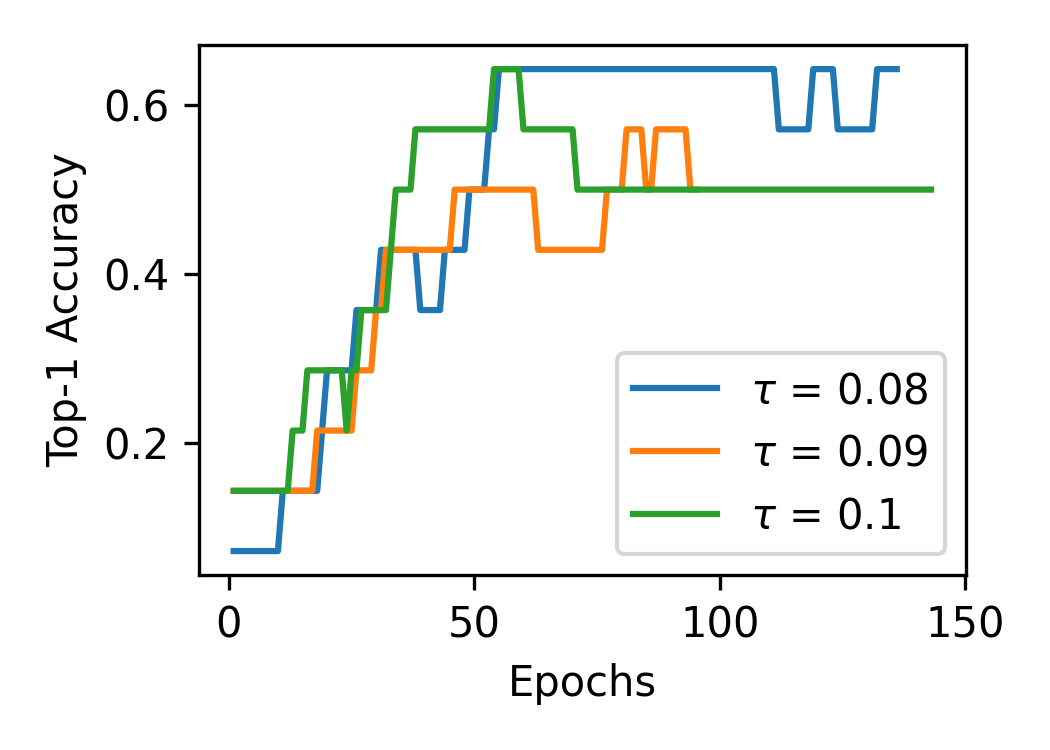}
         \caption{$\tau$}
     \end{subfigure}     
     \begin{subfigure}[b]{.44\linewidth}
         \includegraphics[width=\linewidth]{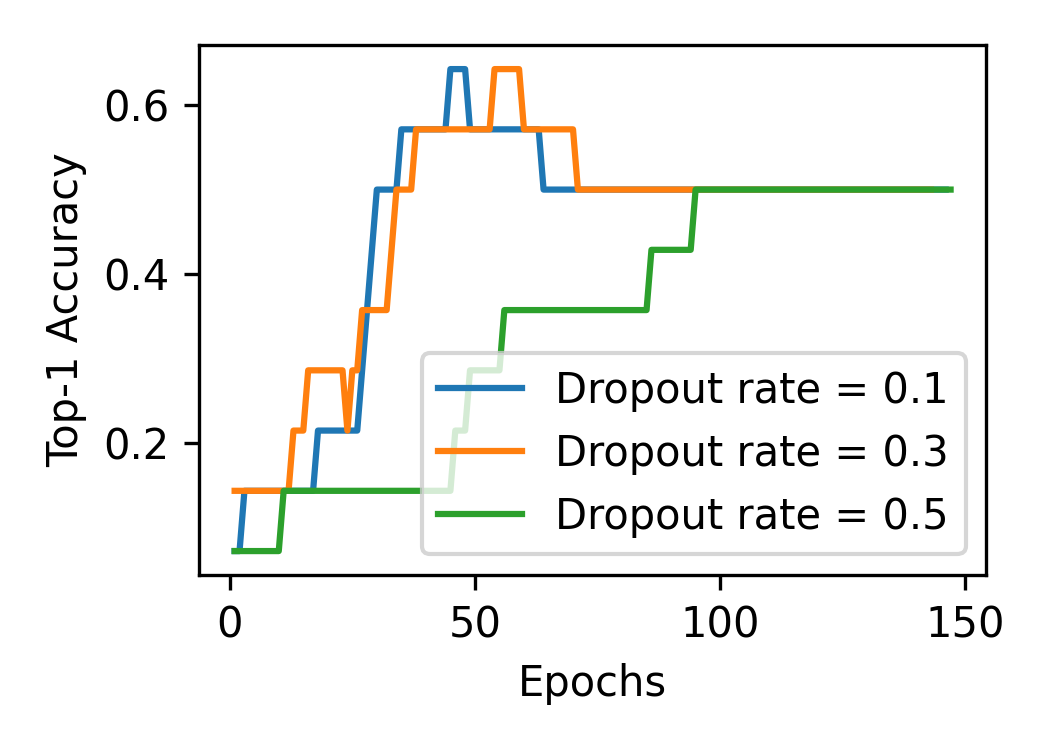}
         \caption{Dropout rate}
     \end{subfigure}
     \caption{Top-1 accuracy on the test set during the fine-tuning of embeddings obtained using LABSE-14, with GPT-4 used for idiomatic meaning generation.}
     \label{fig:params_GPT4}

    \centering
    \begin{subfigure}[b]{.44\linewidth}
         \includegraphics[width=\linewidth]{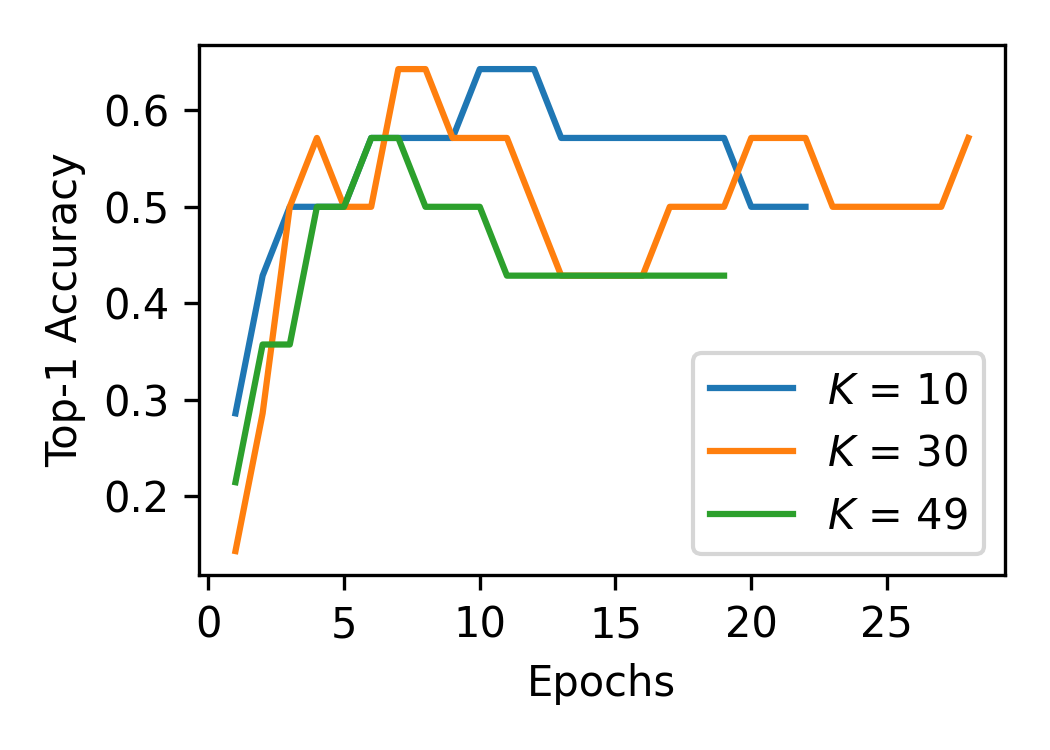} \caption{$K$}
     \end{subfigure}      
     \begin{subfigure}[b]{.44\linewidth}
         \includegraphics[width=\linewidth]{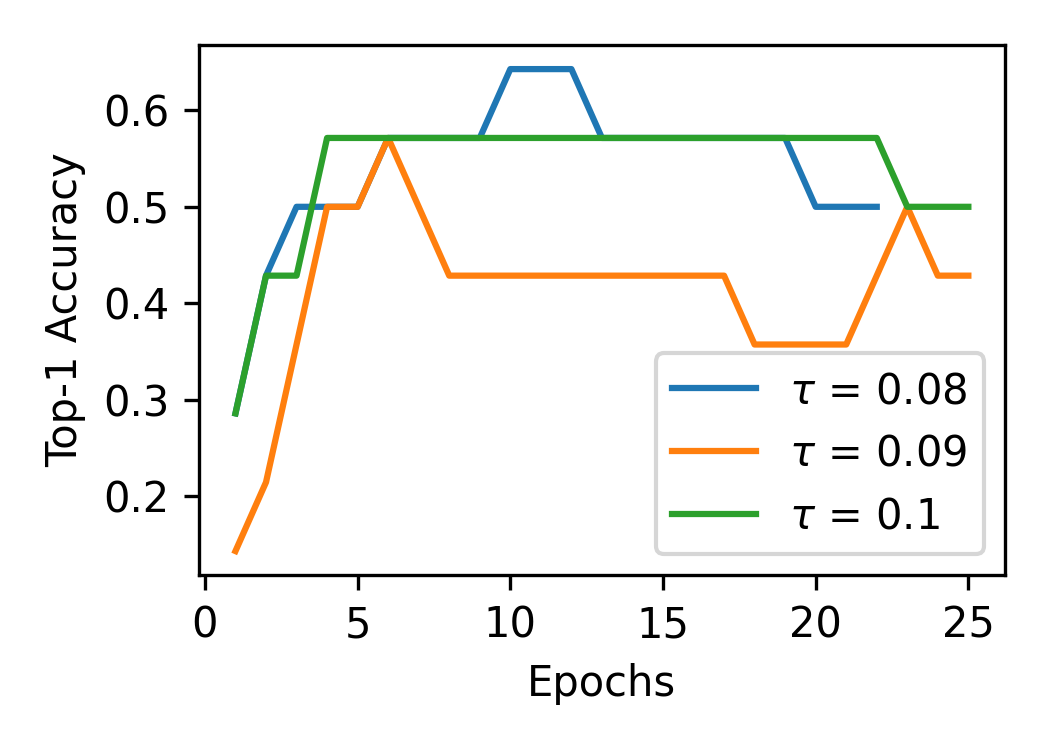}
         \caption{$\tau$}
     \end{subfigure}     
     \begin{subfigure}[b]{.44\linewidth}
         \includegraphics[width=\linewidth]{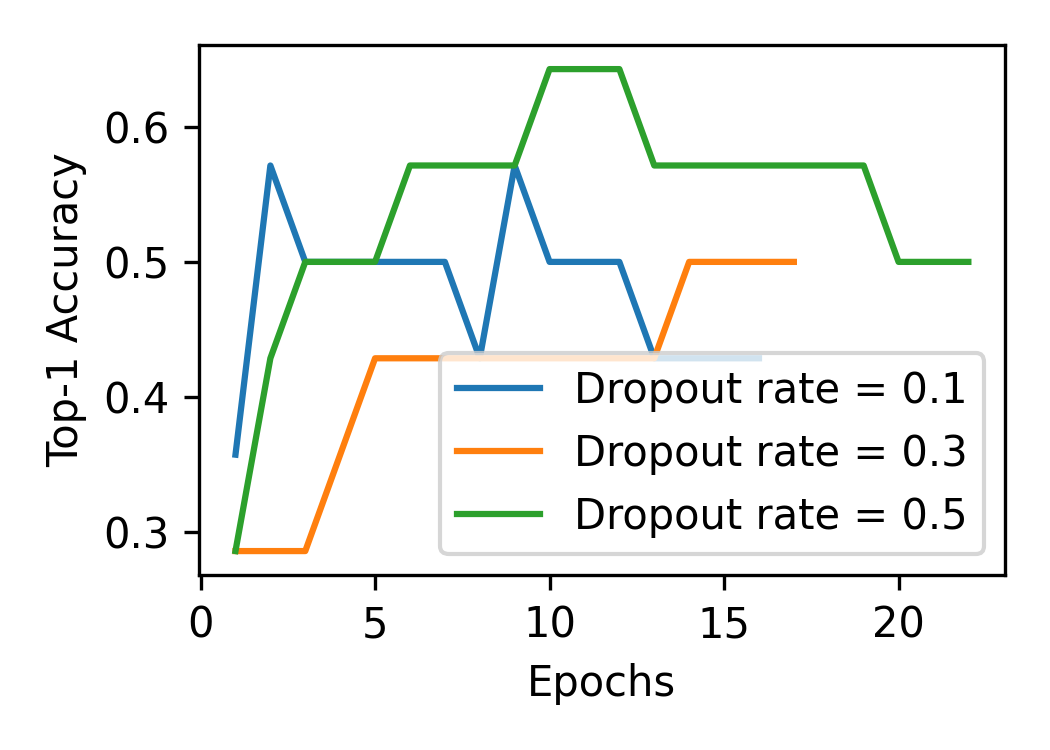}
         \caption{Dropout rate}
     \end{subfigure}
     \caption{Top-1 accuracy on the test set during the fine-tuning of embeddings obtained using LABSE-14, with GPT-4o used for idiomatic meaning generation.}
     \label{fig:params_GPT4o}
\end{figure}

\subsection{Hyperparameter Analysis}

This section explores the impact of key hyperparameters on model fine-tuning performance, including the number of soft negatives ($K$), temperature ($\tau$), and dropout rate. For each fine-tuned model, one hyperparameter was varied while the others remained at their optimal values (as in Table \ref{tab:params_contrast}). Top-1 accuracy was evaluated across training epochs to assess each hyperparameter.
Figures~\ref{fig:params_GPT35}, \ref{fig:params_GPT4}, and \ref{fig:params_GPT4o} illustrate the effect and sensitivity of three hyperparameters on the top-1 accuracy during the fine-tuning of LABSE-14 embeddings, using different GPT variants for idiomatic meaning generation.
Across all models, $K$ exhibited moderate sensitivity, with higher $K$ generally yielding better results for GPT-3.5 and GPT-4. Meanwhile, for GPT-4o, lower $K$ generally performed better.
Temperature $\tau$ showed high sensitivity, with small variations (from 0.08 to 0.1) leading to notable shifts in accuracy. For GPT-3.5 and GPT-4, $\tau = 0.1$ yielded the best results, whereas for GPT-4o, a lower value of $\tau = 0.08$ was optimal.
The dropout rate exhibited model-specific effects. A rate of $0.1$ worked best for GPT-3.5. For GPT-4, lower rates of $0.1$ and $0.3$ yielded similar performance. For both GPT-3.5 and GPT-4, higher dropout rates appeared to degrade performance. In contrast, GPT-4o benefited from a higher rate of $0.5$.

\color{black}


\section{Conclusions}

This work explored the use of generative LLMs and multilingual CLIP models to enhance idiomatic compound representations for image ranking in SemEval-2025 Task 1. By using LLMs to generate idiomatic meanings and leveraging multilingual CLIP models to extract multimodal embeddings, the proposed method improved representation quality compared to using original nominal compounds. Experimental results demonstrated the effectiveness of the proposed method.
For English, the ensemble method using GPT-3.5, GPT-4, and GPT-4o, with the XLM-R Large ViT-B/32 multilingual CLIP model achieved superior performance compared to the other selected LLMs and CLIP models. For Brazilian Portuguese, GPT-3.5 with the LABSE ViT-L/14 multilingual CLIP model outperformed the others.
Fine-tuning CLIP embeddings performed worse than using embeddings extracted from pretrained CLIP models. This is likely due to limitations in fine-tuning data and the capacity of the proposed contrastive learning model. However, it could still be a promising approach for further improvement. Future work could focus on {improving caption utilization (e.g., through different truncation methods and paraphrasing)}, refining fine-tuning strategies and expanding training data to further enhance idiomaticity representation.

\bibliography{task1-biblio}

\begin{thebibliography}{12}
\expandafter\ifx\csname natexlab\endcsname\relax\def\natexlab#1{#1}\fi

\bibitem[{Carlsson et~al.(2022)Carlsson, Eisen, Rekathati, and Sahlgren}]{carlsson-EtAl:2022:LREC}
Fredrik Carlsson, Philipp Eisen, Faton Rekathati, and Magnus Sahlgren. 2022.
\newblock \href {https://aclanthology.org/2022.lrec-1.739} {Cross-lingual and multilingual clip}.
\newblock In \emph{Proceedings of the Language Resources and Evaluation Conference}. European Language Resources Association.

\bibitem[{He et~al.(2024)He, Idiart, Scarton, and Villavicencio}]{he2024enhancing}
Wei He, Marco Idiart, Carolina Scarton, and Aline Villavicencio. 2024.
\newblock Enhancing idiomatic representation in multiple languages via an adaptive contrastive triplet loss.
\newblock \emph{arXiv preprint arXiv:2406.15175}.

\bibitem[{Markchom et~al.(2022)Markchom, Liang, and Chen}]{markchom-etal-2022-uor}
Thanet Markchom, Huizhi Liang, and Jiaoyan Chen. 2022.
\newblock \href {https://doi.org/10.18653/v1/2022.semeval-1.33} {{U}o{R}-{NCL} at {S}em{E}val-2022 task 3: Fine-tuning the {BERT}-based models for validating taxonomic relations}.
\newblock In \emph{Proceedings of the 16th International Workshop on Semantic Evaluation (SemEval-2022)}, pages 260--265, Seattle, United States. Association for Computational Linguistics.

\bibitem[{Oord et~al.(2018)Oord, Li, and Vinyals}]{oord2018representation}
Aaron van~den Oord, Yazhe Li, and Oriol Vinyals. 2018.
\newblock Representation learning with contrastive predictive coding.
\newblock \emph{arXiv preprint arXiv:1807.03748}.

\bibitem[{OpenAI(2024)}]{openaiGPT4o}
OpenAI. 2024.
\newblock \href {hhttps://openai.com/index/hello-gpt-4o/} {Hello {GPT}-4o}.

\bibitem[{Phelps et~al.(2024)Phelps, Pickard, Mi, Gow-Smith, and Villavicencio}]{phelps-etal-2024-sign}
Dylan Phelps, Thomas M.~R. Pickard, Maggie Mi, Edward Gow-Smith, and Aline Villavicencio. 2024.
\newblock Sign of the times: Evaluating the use of large language models for idiomaticity detection.
\newblock In \emph{Proceedings of the Joint Workshop on Multiword Expressions and Universal Dependencies}, Italia.

\bibitem[{Pickard et~al.(2025)Pickard, Villavicencio, Mi, He, Phelps, Scarton, and Idiart}]{pickard2025semeval}
Thomas Pickard, Aline Villavicencio, Maggie Mi, Wei He, Dylan Phelps, Carolina Scarton, and Marco Idiart. 2025.
\newblock Semeval-2025 task 1: Admire - advancing multimodal idiomaticity representation.
\newblock In \emph{Proceedings of the 19th International Workshop on Semantic Evaluations (SemEval-2025)}, Vienna, Austria. Association for Computational Linguistics.

\bibitem[{Radford et~al.(2021)Radford, Kim, Hallacy, Ramesh, Goh, Agarwal, Sastry, Askell, Mishkin, Clark, Krueger, and Sutskever}]{radford2021learningtransferablevisualmodels}
Alec Radford, Jong~Wook Kim, Chris Hallacy, Aditya Ramesh, Gabriel Goh, Sandhini Agarwal, Girish Sastry, Amanda Askell, Pamela Mishkin, Jack Clark, Gretchen Krueger, and Ilya Sutskever. 2021.
\newblock \href {http://arxiv.org/abs/2103.00020} {Learning transferable visual models from natural language supervision}.

\bibitem[{Radford et~al.(2018)Radford, Narasimhan, Salimans, Sutskever et~al.}]{Radford2018ImprovingLU}
Alec Radford, Karthik Narasimhan, Tim Salimans, Ilya Sutskever, et~al. 2018.
\newblock Improving language understanding by generative pre-training.
\newblock San Francisco, CA, USA.

\bibitem[{Raffel et~al.(2020)Raffel, Shazeer, Roberts, Lee, Narang, Matena, Zhou, Li, and Liu}]{2020t5}
Colin Raffel, Noam Shazeer, Adam Roberts, Katherine Lee, Sharan Narang, Michael Matena, Yanqi Zhou, Wei Li, and Peter~J. Liu. 2020.
\newblock Exploring the limits of transfer learning with a unified text-to-text transformer.
\newblock \emph{Journal of Machine Learning Research}.

\bibitem[{Tiedemann et~al.(2023)Tiedemann, Aulamo, Bakshandaeva, Boggia, Gr{\"o}nroos, Nieminen, Raganato\, Scherrer, Vazquez, and Virpioja}]{tiedemann2023democratizing}
J{\"o}rg Tiedemann, Mikko Aulamo, Daria Bakshandaeva, Michele Boggia, Stig-Arne Gr{\"o}nroos, Tommi Nieminen, Alessandro Raganato\, Yves Scherrer, Raul Vazquez, and Sami Virpioja. 2023.
\newblock \href {https://doi.org/10.1007/s10579-023-09704-w} {Democratizing neural machine translation with {OPUS-MT}}.
\newblock \emph{Language Resources and Evaluation}, pages 713--755.

\bibitem[{Xiong et~al.(2024)Xiong, Markchom, Zheng, Jung, Ojha, and Liang}]{xiong-etal-2024-ncl}
Feng Xiong, Thanet Markchom, Ziwei Zheng, Subin Jung, Varun Ojha, and Huizhi Liang. 2024.
\newblock {NCL}-{U}o{R} at {S}em{E}val-2024 task 8: Fine-tuning large language models for multigenerator, multidomain, and multilingual machine-generated text detection.
\newblock In \emph{Proceedings of the 18th International Workshop on Semantic Evaluation}, Mexico.

\end{thebibliography}

\appendix

\section{Hyperparameters Settings}
\label{apd:hyperparams}

Table \ref{tab:params_contrast} shows the selected hyperparameters for different contrastive learning models in the \textbf{CI-F} and \textbf{CIC-F} approaches.

\begin{table}[ht]
\caption{The selected hyperparameters for different contrastive learning models.}
    \centering
    \scalebox{.7}{
    \begin{tabular}{lp{1cm}p{1.2cm}llp{1.25cm}}
    \toprule
        \bf Model         & \bf Batch Size & \bf Learning Rate & $\boldsymbol K$ & \bf $\boldsymbol \tau$ & \bf Dropout Rate \\ \midrule
        GPT-3.5 + LABSE-14      & 16         & 1e-5          & 49            & 0.1         & 0.1  \\
        GPT-4 + LABSE-14        & 16         & 1e-5          & 30            & 0.1         & 0.3  \\
        GPT-4o + LABSE-14       & 16         & 1e-4          & 10            & 0.08        & 0.5  \\
        \bottomrule
    \end{tabular}}
    \label{tab:params_contrast}
\end{table}

\section{Detailed Evaluation Results}
\label{apd:detailed_eval_results}

Table \ref{tab:dev_results} shows the results on English and Portuguese development sets.

\begin{table}[ht]

    \caption{Evaluation results for English (EN) and Portuguese (PT) development sets, with the highest values in bold and the second-highest underlined.}
    
    \centering
    \scalebox{.6}{
    \begin{tabular}{llrrrrrr}
    \toprule
    \multirow{2}{*}{\textbf{LLM}} & \multirow{2}{*}{\textbf{CLIP model}} & \multicolumn{3}{c}{\textbf{Dev EN}} & \multicolumn{3}{c}{\textbf{Dev PT}} \\ \cmidrule(lr){3-5} \cmidrule(lr){6-8}
     &  & \multicolumn{1}{l}{\textbf{Acc}} & \multicolumn{1}{l}{\textbf{Corr}} & \multicolumn{1}{l}{\textbf{DCG}} & \multicolumn{1}{l}{\textbf{Acc}} & \multicolumn{1}{l}{\textbf{Corr}} & \multicolumn{1}{l}{\textbf{DCG}} \\ \midrule
    \multicolumn{8}{l}{\cellcolor[HTML]{C0C0C0}Use only compound and image embeddings without fine-tuning} \\
    
    GPT-3.5 & XLM-14 & 0.600 & 0.313 & 3.055 & \textbf{0.400} & \textbf{0.320} & \textbf{2.620} \\
    GPT-4 & XLM-14 & 0.533 & 0.193 & 2.818 & \textbf{0.400} & 0.220 & 2.562 \\
    GPT-4o & XLM-14 & 0.600 & 0.233 & 2.943 & \textbf{0.400} & 0.220 & {\ul 2.582} \\
    Ensemble & XLM-14 & 0.600 & {\ul 0.353} & 3.005 & \textbf{0.400} & 0.260 & {\ul 2.582} \\
    GPT-3.5 & XLM-32 & \textbf{0.733} & \textbf{0.427} & \textbf{3.219} & \textbf{0.400} & 0.160 & 2.487 \\
    GPT-4 & XLM-32 & 0.533 & 0.273 & 2.794 & {\ul 0.300} & 0.230 & 2.375 \\
    GPT-4o & XLM-32 & 0.600 & 0.293 & 2.918 & {\ul 0.300} & 0.050 & 2.338 \\
    Ensemble & XLM-32 & {\ul 0.667} & 0.260 & 3.005 & {\ul 0.300} & 0.110 & 2.375 \\
    GPT-3.5 & LABSE-14 & 0.600 & 0.293 & 3.006 & 0.200 & {\ul 0.280} & 2.376 \\
    GPT-4 & LABSE-14 & 0.533 & 0.153 & 2.781 & {\ul 0.300} & {\ul 0.280} & 2.469 \\
    GPT-4o & LABSE-14 & 0.600 & 0.153 & 2.993 & {\ul 0.300} & {\ul 0.280} & 2.413 \\
    Ensemble & LABSE-14 & 0.600 & 0.253 & 2.919 & {\ul 0.300} & 0.240 & 2.450 \\ \midrule
    \multicolumn{8}{l}{\cellcolor[HTML]{C0C0C0}Use compound image and caption embeddings without fine-tuning} \\
    GPT-3.5 & XLM-14 & 0.400 & 0.013 & 2.682 & {\ul 0.300} & -0.120 & 2.452 \\
    GPT-4 & XLM-14 & 0.400 & 0.040 & 2.645 & {\ul 0.300} & -0.030 & 2.452 \\
    GPT-4o & XLM-14 & 0.467 & 0.273 & 2.719 & {\ul 0.300} & -0.080 & 2.452 \\
    Ensemble & XLM-14 & 0.400 & 0.087 & 2.682 & {\ul 0.300} & -0.100 & 2.452 \\
    GPT-3.5 & XLM-32 & 0.533 & 0.240 & 2.970 & {\ul 0.300} & -0.070 & 2.508 \\
    GPT-4 & XLM-32 & 0.467 & 0.173 & 2.719 & {\ul 0.300} & -0.030 & 2.508 \\
    GPT-4o & XLM-32 & 0.533 & 0.267 & 2.857 & 0.200 & -0.020 & 2.396 \\
    Ensemble & XLM-32 & 0.533 & 0.260 & 2.794 & {\ul 0.300} & -0.050 & 2.508 \\
    GPT-3.5 & LABSE-14 & 0.400 & -0.140 & 2.671 & 0.200 & -0.210 & 2.378 \\
    GPT-4 & LABSE-14 & 0.467 & -0.020 & 2.707 & 0.200 & -0.130 & 2.378 \\
    GPT-4o & LABSE-14 & 0.467 & -0.067 & 2.682 & 0.200 & -0.190 & 2.378 \\
    Ensemble & LABSE-14 & 0.400 & -0.013 & 2.620 & 0.200 & -0.170 & 2.378 \\ \midrule
    \multicolumn{8}{l}{\cellcolor[HTML]{C0C0C0}Use only compound and image embeddings with fine-tuning} \\
    GPT-3.5 & LABSE-14 & 0.600 & 0.213 & {\ul 3.159} & - & - & - \\
    GPT-4 & LABSE-14 & 0.600 & 0.107 & 3.019 & - & - & - \\
    GPT-4o & LABSE-14 & {\ul 0.667} & 0.187 & 3.131 & - & - & - \\ \midrule
    \multicolumn{8}{l}{\cellcolor[HTML]{C0C0C0}Use compound image and caption embeddings with fine-tuning} \\
    GPT-3.5 & LABSE-14 & 0.600 & 0.127 & 3.158 & - & - & - \\
    GPT-4 & LABSE-14 & 0.533 & 0.047 & 2.844 & - & - & - \\
    GPT-4o & LABSE-14 & 0.600 & 0.113 & 3.005 & - & - & - \\ \bottomrule

    \end{tabular}}

    \label{tab:dev_results}
    
\end{table}

\begin{table*}[ht]

    \caption{Evaluation results on the English (EN), Portuguese (PT), Extended English (XE) and Extended Portuguese (XP) test sets. The highest values in each column are in bold, and the second-highest values are underlined.}

    \centering
    \scalebox{.75}{
    \begin{tabular}{llllllllllllll}
    \toprule

    \multirow{2}{*}{\textbf{LLM}} & \multirow{2}{*}{\textbf{CLIP model}} & \multicolumn{3}{c}{\textbf{Test EN}} & \multicolumn{3}{c}{\textbf{Test PT}} & \multicolumn{3}{c}{\textbf{Test XE}} & \multicolumn{3}{c}{\textbf{Test XP}} \\

    \cmidrule(lr){3-5} \cmidrule(lr){6-8} \cmidrule(lr){9-11} \cmidrule(lr){12-14}
    
     &  & \textbf{Acc} & \textbf{Corr} & \textbf{DCG} & \textbf{Acc} & \textbf{Corr} & \textbf{DCG} & \textbf{Acc} & \textbf{Corr} & \textbf{DCG} & \textbf{Acc} & \textbf{Corr} & \textbf{DCG} \\
 
     \midrule
    \multicolumn{14}{l}{\cellcolor[HTML]{C0C0C0}Baselines} \\
    - & XLM-14 & 0.333 & -0.027 & 2.579 & 0.385 & \bf 0.415 & 2.661 & 0.410 & 0.062 & 2.671 & 0.345 & 0.087 & 2.573 \\
    - & XLM-32 & 0.267 & -0.173 & 2.482 & 0.385 & 0.223 & 2.669 & 0.440 & 0.087 & 2.681 & 0.400 & 0.102 & 2.646 \\
    - & LABSE-14 & 0.467 & 0.120 & 2.706 & 0.385 & 0.146 & 2.578 & 0.400 & 0.127 & 2.639 & 0.455 & 0.164 & 2.680 \\

    \midrule
    \multicolumn{14}{l}{\cellcolor[HTML]{C0C0C0}Use only compound and image embeddings without fine-tuning} \\
    GPT-3.5 & XLM-14 & 0.533 & 0.220 & 2.943 & 0.385 & \bf 0.415 & 2.637 & 0.470 & 0.157 & 2.815 & 0.436 & 0.047 & 2.746 \\
    GPT-4 & XLM-14 & 0.533 & 0.133 & 2.970 & \bf 0.538 & \underline{0.354} & 2.951 & 0.500 & 0.125 & 2.897 & 0.400 & 0.127 & 2.719 \\
    GPT-4o & XLM-14 & 0.467 & 0.193 & 2.867 & \bf 0.538 & 0.285 & \bf 3.045 & 0.480 & 0.093 & 2.903 & 0.400 & 0.115 & 2.724 \\
    Ensemble & XLM-14 & 0.533 & 0.233 & 2.921 & \underline{0.462} & 0.269 & 2.792 & 0.510 & 0.130 & \underline{2.919} & 0.364 & 0.151 & 2.722 \\
    GPT-3.5 & XLM-32 & 0.333 & -0.013 & 2.690 & \underline{0.462} & 0.131 & 2.749 & 0.450 & 0.119 & 2.767 & \underline{0.491} & 0.104 & 2.834 \\
    GPT-4 & XLM-32 & 0.533 & 0.167 & 2.940 & 0.385 & 0.223 & 2.747 & \underline{0.540} & \underline{0.186} & 2.891 & 0.418 & 0.125 & 2.729 \\
    GPT-4o & XLM-32 & 0.467 & 0.087 & 2.849 & \bf 0.538 & 0.092 & \underline{2.953} & 0.520 & 0.184 & 2.907 & 0.473 & 0.184 & 2.810 \\
    Ensemble & XLM-32 & 0.467 & 0.053 & 2.821 & \bf 0.538 & 0.169 & 2.866 & \bf 0.570 & 0.132 & \bf 2.957 & 0.436 & 0.176 & 2.788 \\
    GPT-3.5 & LABSE-14 & \bf 0.667 & \bf 0.360 & \bf 3.102 & 0.308 & 0.123 & 2.486 & 0.440 & 0.149 & 2.773 & \bf 0.564 & \bf 0.193 & \bf 2.900 \\
    GPT-4 & LABSE-14 & \underline{0.600} & 0.147 & \underline{2.993} & \underline{0.462} & 0.131 & 2.771 & 0.480 & 0.165 & 2.867 & 0.473 & 0.185 & 2.779 \\
    GPT-4o & LABSE-14 & 0.533 & \underline{0.267} & 2.963 & \bf 0.538 & 0.223 & 2.947 & 0.500 & 0.175 & 2.889 & 0.473 & \underline{0.189} & 2.807 \\
    Ensemble & LABSE-14 & \underline{0.600} & 0.247 & 2.985 & \underline{0.462} & 0.269 & 2.691 & 0.510 & \bf 0.187 & 2.902 & \underline{0.491} & 0.187 & \underline{2.852} \\

    \midrule
    \multicolumn{14}{l}{\cellcolor[HTML]{C0C0C0}Use compound image and caption embeddings without fine-tuning} \\
    GPT-3.5 & XLM-14 & 0.333 & 0.087 & 2.566 & 0.231 & 0.023 & 2.337 & 0.280 & 0.037 & 2.468 & 0.327 & 0.002 & 2.527 \\
    GPT-4 & XLM-14 & 0.400 & 0.153 & 2.645 & 0.154 & 0.054 & 2.278 & 0.280 & 0.037 & 2.468 & 0.327 & 0.002 & 2.527 \\
    GPT-4o & XLM-14 & 0.333 & 0.153 & 2.888 & 0.231 & 0.046 & 2.378 & 0.290 & 0.050 & 2.525 & 0.327 & 0.020 & 2.553 \\
    Ensemble & XLM-14 & 0.333 & 0.113 & 2.566 & 0.308 & 0.092 & 2.433 & 0.280 & 0.053 & 2.501 & 0.291 & 0.067 & 2.498 \\
    GPT-3.5 & XLM-32 & 0.267 & 0.060 & 2.456 & 0.154 & 0.223 & 2.344 & 0.320 & 0.049 & 2.563 & 0.418 & 0.118 & 2.675 \\
    GPT-4 & XLM-32 & 0.333 & 0.047 & 2.527 & 0.154 & 0.215 & 2.344 & 0.360 & 0.078 & 2.603 & 0.400 & 0.091 & 2.665 \\
    GPT-4o & XLM-32 & 0.267 & 0.127 & 2.456 & 0.154 & 0.262 & 2.354 & 0.380 & 0.104 & 2.680 & 0.418 & 0.038 & 2.695 \\
    Ensemble & XLM-32 & 0.267 & 0.020 & 2.448 & 0.154 & 0.223 & 2.344 & 0.380 & 0.034 & 2.653 & 0.418 & 0.044 & 2.685 \\
    GPT-3.5 & LABSE-14 & 0.333 & 0.027 & 2.569 & 0.308 & -0.008 & 2.440 & 0.240 & 0.047 & 2.450 & 0.291 & 0.069 & 2.526 \\
    GPT-4 & LABSE-14 & 0.333 & 0.007 & 2.562 & 0.308 & 0.023 & 2.480 & 0.270 & 0.073 & 2.520 & 0.273 & 0.098 & 2.477 \\
    GPT-4o & LABSE-14 & 0.333 & 0.027 & 2.569 & 0.308 & 0.077 & 2.530 & 0.330 & 0.079 & 2.594 & 0.273 & 0.076 & 2.497 \\
    Ensemble & LABSE-14 & 0.333 & -0.020 & 2.567 & 0.308 & 0.054 & 2.496 & 0.270 & 0.079 & 2.519 & 0.291 & 0.095 & 2.502 \\

    \midrule
    \multicolumn{14}{l}{\cellcolor[HTML]{C0C0C0}Use only compound and image embeddings with fine-tuning} \\
    GPT-3.5 & LABSE-14 & 0.400 & 0.107 & 2.814 &  -&  - &  - & 0.390 & 0.015 & 2.694 &  - &  - &  - \\
    GPT-4 & LABSE-14 & 0.333 & 0.233 & 2.784 &  - &  - &  - & 0.410 & 0.056 & 2.777 &  - &  - &  - \\
    GPT-4o & LABSE-14 & 0.267 & -0.073 & 2.676 &  - &  - &  - & 0.380 & 0.075 & 2.711 &  - &  - &  - \\

    \midrule
    \multicolumn{14}{l}{\cellcolor[HTML]{C0C0C0}Use compound image and caption embeddings with fine-tuning} \\
    GPT-3.5 & LABSE-14 & 0.333 & 0.051 & 2.719 & - & - & - & 0.400 & 0.053 & 2.694 & - & - & - \\
    GPT-4 & LABSE-14 & 0.267 & 0.133 & 2.724 & - & - & - & 0.440 & 0.158 & 2.826 & - & - & - \\
    GPT-4o & LABSE-14 & 0.267 & 0.040 & 2.607 & - & - & - & 0.390 & 0.091 & 2.740 & - & - & -\\

    \bottomrule
    \end{tabular}}

    \label{tab:appendix_eval_results}
\end{table*}

Table \ref{tab:appendix_eval_results} shows detailed evaluation results for the baselines and variations of proposed method on the English (EN), Portuguese (PT), Extended English (XE), and Extended Portuguese (XP) test sets.

\end{document}